# A Deep Q-Learning Agent for the L-Game with Variable Batch Training

Petros Giannakopoulos and Yannis Cotronis

National and Kapodistrian University of Athens - Dept of Informatics and Telecommunications
Ilisia, 15784 - Greece

**Abstract.** We employ the Deep Q-Learning algorithm with Experience Replay to train an agent capable of achieving a high-level of play in the L-Game while self-learning from low-dimensional states. We also employ variable batch size for training in order to mitigate the loss of the rare reward signal and significantly accelerate training. Despite the large action space due to the number of possible moves, the low-dimensional state space and the rarity of rewards, which only come at the end of a game, DQL is successful in training an agent capable of strong play without the use of any search methods or domain knowledge.

## 1 Introduction

### 1.1 Related Work

The seminal Deep-Mind's paper [1] demonstrated how to make deep reinforcement learning obtain human-level performance on a large set of Atari 2600 games. Their method, called Deep Q-Learning involved Q-learning, a deep convolutional neural network followed by one fully connected hidden layer and one fully connected output layer, and a set of additional techniques. The input to the learning system involved high-dimensional visual pixel data.

Deep-Mind later introduced AlphaGo [2], a Go playing program that combines Monte Carlo tree search with convolutional neural networks for searching (policy network) and evaluating (value network) positions, with Deep Reinforcement Learning used for training both networks and supervised mentoring, used in the initial stages of training, with games pulled from a database of games between human players.

An agent for the game of Hex, called NeuroHex was recently introduced [3]. The system for training the agent employed Deep Q-Learning and consisted of a convolutional network followed by one fully connected output layer, differing from AlphaGo's architecture which was fully convolutional. They also used supervised mentoring for network initialization.

[4] showed that DQL can also be effective for reinforcement learning problems with low-dimensional states. They achieved state-of-the-art performance in Keepaway soccer using a shallow network made only from fully connected layers and trained directly from the low-dimensional input state.

One of the earliest and most successful examples of applying Reinforcement Learning combined with a neural network to board games is TD-gammon [5]. There, the self-trained agent achieved superhuman levels of play by approximating state-action values using the Temporal Difference learning method.

## 1.2 The L-Game

The L-Game is an abstract strategy board game created by Edward de Bono and was first presented in his book [6]. His goal was to create a game that, contrary to games like Chess which achieved difficulty through complexity, would be simple yet would still require a high degree of skill.

The game is played by two players. The board consists of sixteen squares (4x4). Each player has a flat 3x2 L-shaped piece that exactly covers four squares. In addition, there are two circular neutral pieces, each occupying a 1x1 square. The starting position is shown in Figure 1.

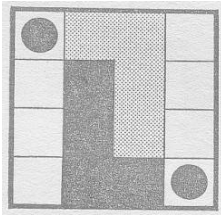
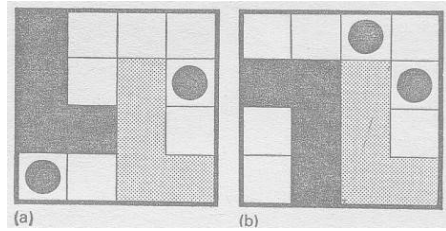

Fig. 1: The L-Game's board and starting positions.

Fig. 2: One of the final winning arrangements in the L-Game.

On each turn, a player picks up his L-piece and replaces it on the board to cover four empty squares, at least one of which must be different from the four squares just vacated. The piece may be rotated or even flipped over if desired. After the L-piece has been placed, the player may optionally choose one (but not both) of the neutral pieces and move it to any vacant square. The objective of the game is to leave your opponent without a legal move for his L-piece. In Figure 2(a) it is Black's turn to play. He first repositions his L-piece and then a neutral piece as shown in 2(b). Now White is unable to move his L-piece and thus loses.

The L-Game is a perfect information, zero-sum game. Neither player has an advantage in any way (e.g. by starting first). In a game with two perfect players, neither will ever win nor lose. The state size of the L-Game allows for precise analysis. There are known to be 2296 different possible valid ways the pieces can be arranged, not counting a rotation or mirror of an arrangement as a new arrangement (which would bring the total arrangements up to 18368) and considering the two neutral pieces to be identical. Any arrangement can be reached during the game, with it being any player's turn. There are 15 basic winning positions, where one of the L-pieces is blocked. Another 14 positions are known to lead to a win after a maximum of 5 moves. Building a winning strategy requires memory of the known winning positions, spatial acuity, and the ability to plan ahead.

The domain can be modelled as a fully observable Markov Decision Process (MDP) where, for each player, a state $s$ is represented by the position of the player and neutral pieces on the game board. Action $a$ is a legal move chosen by the player which leads to the new state of the game board $s'$. If $a$ ends the game, $s'$ is a terminal state and each player receives a reward $r$, positive or negative, depending if he won or lost the game.

## 2   Overview of This Work

### 2.1   Challenges

In this work we explore the application of Reinforcement Learning [7] with Deep Q-Learning to the L-Game. There are some additional challenges involved in applying this method, so successful with Atari, to the L-Game and to other board games with similar general domain characteristics.

One challenge is the large number of actions: up to 128 possible moves for each state of the L-Game. Since Q-learning performs a maximization over all available actions, this large number might cause the noise in estimation to overwhelm the useful signal, resulting in catastrophic maximization bias. Despite this, we found that the linear network was still able to achieve good learning results from the low-dimensional input states.

Another challenge is that the reward signal occurs only at the end of a game, so it is infrequent, coming after a sequence of several turns. This means that most updates are based only on network evaluations without immediate win/loss feedback. The question is whether the learning process will allow this end-of-game reward information to propagate back to the middle and early game. To address this challenge, we save and then assemble all the states for a single played game, from the first state to the last state (were the reward signal appears), into a batch. We then use this batch to update the gradient. Essentially, we perform batch updates but the batch size is not fixed but variable and equal to the number of states contained in a single played game. This helps to minimize, as much as possible, the variance of stochastic gradient updates and allows reaching good results with fewer training epochs (games).

### 2.2   Problem Definition

We use a feedback scheme where a win is set to be worth a reward of +1 and a loss a reward of -1. The reward given for all intermediate moves between the start and end of game is 0. Therefore, the ground truth for the Q-value of every possible state-action pair *q(S,A)* assumes a binary form with its value being either 1 or -1 for every possible state-action pair. We want the agent to maximize his reward during a single game (or episode) by winning. The network approximates the Q-value of every action for a given state, or simply the subjective probability that a particular move will result in a win minus the probability that it will result in a loss. The accurate prediction of the true value of *q(S,A)* for all states encountered during play is the measure of how strong a player the agent will be by following the policy π which takes the action *a* with the highest estimated Q-value (or win probability) *argmax[Q(s)[a]]* for each state *s*.

### 2.3   Problem Modelling

We use the Torch library [8] to build and train out network. The input state size is the size of the board: 4x4 = 16 data points. This dictates the use of an input layer to the network consisting of 16 nodes. There up to 128 possible moves (or actions) which points to the use of an output layer consisting of 128 nodes. We experimented with various sizes and number of hidden layers and we found that the best results were obtained with just 2 hidden layers of 512 nodes width each. Due to the low dimensionality of the input state, deep architectures do not give an advantage in the learning process since there's less need to model complex non-linear relationships in problems with simple input such in this case. Activation function for every layer except the output was selected to be Rectified Linear Units (ReLU). The output of the network is a vector containing the value of every possible move corresponding to the input state. Illegal moves (occupied positions) are still evaluated by the network but are subsequently pruned.

### 2.4 Learning Process and Results

We do not use any mentoring to initialize the network. All the learning is performed through self-play. We found that random exploration during training was enough for the network to experience and successfully learn from a wide variety of game positions.

We implement Deep Q-learning with Experience Replay as first introduced in [1]. Our implementation differentiates in the fact that an episode (=game) consists of a sequence of state-action pairs with a singular reward being given at the end of the game based on win or loss and no rewards being given for actions in-between the start and the end. For this reason, storing and randomly sampling state-action pairs from the replay memory either individually or in small batches may cause the rare reward signal to be lost in the estimation noise. In an attempt to mitigate this, we temporarily keep all the state-action pairs encountered during a game in a table and, once the game ends, we push this game and all the experiences acquired with it in the experience replay memory as shown in Figure 3. Sampling from the replay memory is done in similar fashion: a full previously played game is randomly sampled and used for batch-updating the weights of the network.

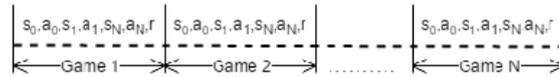

Fig. 3: Structure of the Replay Memory. Each game consists of an arbitrary amount of state-action pairs leading to the final state and the reward.

We save a large set of the most recently played games (10000) in the Replay Memory and sample N games from that set on each recall from the memory. A value of N = 0 means that Replay Memory is not used and it serves as the baseline. We experimented with the impact the number of games sampled per recall has on learning performance for the first 25000 epochs (Figure 4). It is evident that the use of experience replay improves performance considerably over baseline up to N = 10 games sampled per recall. Higher sample sizes do not appear to provide additional benefits and very high sample sizes may even slow down learning. We ended up using a sample size of N = 10 games per recall.

Since we want to place as much weight as possible to the outcome of a game versus intermediate moves, we need to use a high discount factor $\gamma$. We tried values for $\gamma$ between 0.7 and 1 and we found a value of $\gamma = 0.9$ to deliver the best training results. Lower values yielded slower convergence while higher did not provide any further acceleration of learning and values approaching 1 caused action values to diverge.

We used an annealed ϵ schedule for the ϵ-greedy policy. We decrease it linearly from 0.05 to 0.01, meaning the agent makes a random move from 5% of the time at the beginning of training to 1% near the end. ϵ is fixed at 0.01 during validating.

For the backpropagation algorithm, we experimented with ADADELTA [9], RMSProp and SGD with Nesterov momentum [10]. We found SGD with Nesterov momentum and an annealed learning rate schedule to deliver the best results, achieving faster approach to convergence and the lowest final error. The performance of each backpropagation algorithm for the first 25000 epochs is shown in Figure 5.

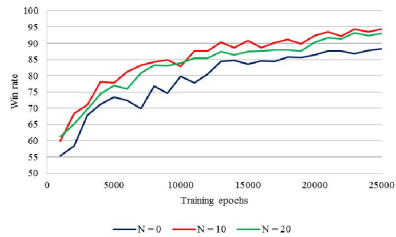
Fig. 4: Effect of Replay Memory sample size on learning performance.

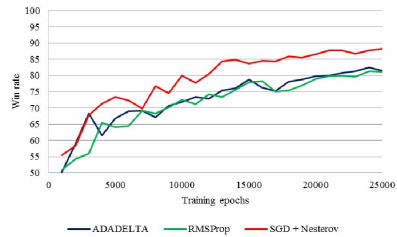
Fig. 5: Performance of the backpropagation algorithms tested.

Figure 6 shows the learning performance for the first 25000 epochs when using a fixed mini-batch size of 32 episodes [(s, a, r, s') pairs] per learning step to update the gradients, vs. using an entire game consisting of an arbitrary number of episodes. For training, the agent assumes the roles of both players and begins learning through self-play for 100000 games. To gauge the agent's progress we periodically (every 1000 training episodes) validate by playing 1000 games against an agent playing randomly and against a perfect agent based on Minimax who can never lose. Training took 9 hours on an Intel® Core™ i7-3770K. The final trained agent achieves a 98% winning rate, playing as either Player 1 or Player 2, over 10000 games versus the random player, while making a random move every 100 moves. Furthermore, it achieves a 95% draw rate versus a perfect (undefeatable) minimax player. We consider a game to be drawn in this case, if it goes on for over 100 turns without a winner.

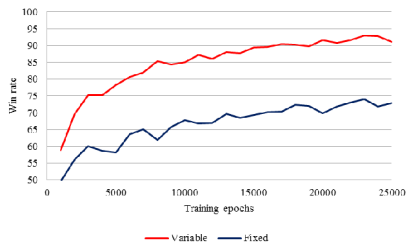
Fig. 6: Learning rate using a fixed mini-batch size vs. using variable.

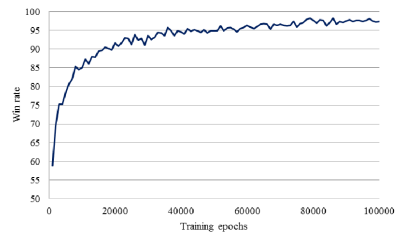
Fig. 7: Agent performance versus a random player as training progresses.

## 2.5 Conclusion

In this paper we developed a game playing agent based on Deep Q-Learning for a challenging board game. We evaluated the performance of Deep Q-Learning on a task which involves, contrary to the original Atari video game playing application of Deep Q-Learning, learning from low-dimensional states, a large action space and a rare singular reward coming at the end of an episode. We also did not use mentoring for guiding the network at the initial stages of training and all exploration is performed by the agent. The results of the experiments show that Deep Q-Learning is able to produce well-performing agents, in spite of these challenges. Our agent achieves playing performance close to a perfect player, without any domain knowledge and mentoring, purely through self-play. Deep Learning techniques do increase performance, with experience replay significantly improving the results. For the backpropagation algorithm, we found SGD with Nesterov momentum and a decaying learning rate to perform better in this task compared to adaptive methods such as RMSProp and ADADELTA. We found that the use of variable batch training can provide substantial benefits to the performance of Reinforcement Learning applications on the domain of board games, where the reward signal is inherently rare, allowing an agent to learn more efficiently from it.